# Automatic Facial Feature Extraction and Expression Recognition based on Neural Network

S.P.Khandait
Deptt of IT, KDKCE
Nagpur, Maharashtra, India
prapti_khandait@yahoo.co.in

Dr. R.C.Thool
Deptt. of IT, SGGSIET
Nanded, Maharashtra, India
rcthool@yahoo.com

P.D.Khandait
Deptt. of Etrx, KDKCE,
Nagpur, Maharashtra, India

*Abstract*— In this paper, an approach to the problem of automatic facial feature extraction from a still frontal posed image and classification and recognition of facial expression and hence emotion and mood of a person is presented. Feed forward back propagation neural network is used as a classifier for classifying the expressions of supplied face into seven basic categories like surprise, neutral, sad, disgust, fear, happy and angry. For face portion segmentation and localization, morphological image processing operations are used. Permanent facial features like eyebrows, eyes, mouth and nose are extracted using SUSAN edge detection operator, facial geometry, edge projection analysis. Experiments are carried out on JAFFE facial expression database and gives better performance in terms of 100% accuracy for training set and 95.26% accuracy for test set.

## I. INTRODUCTION

Due to technological advancements; there is an arousal of the world where human being and intelligent robots live together. Area of Human Computer Interaction (HCI) plays an important role in resolving the absences of neutral sympathy in interaction between human being and machine (computer). HCI will be much more effective and useful if computer can predict about emotional state of human being and hence mood of a person from supplied images on the basis of facial expressions. Mehrabian [1] pointed out that 7% of human communication information is communicated by linguistic language (verbal part), 38% by paralanguage (vocal part) and 55% by facial expression. Therefore facial expressions are the most important information for emotions perception in face to face communication. For classifying facial expressions into different categories, it is necessary to extract important facial features which contribute in identifying proper and particular expressions. Recognition and classification of human facial expression by computer is an important issue to develop automatic facial expression recognition system in vision community. In recent years, much research has been done on machine recognition of human facial expressions [2-6]. In last few years, use of computers for Facial expression and emotion recognition and its related information use in HCI has gained significant research interest which in tern given rise to a number of automatic methods to recognize facial expressions in images or video [7-12]. This paper explains about an approach to the problem of facial feature extraction from a still frontal posed image and classification and recognition of facial expression and hence emotion and mood of a person. Feed forward back propagation neural network is used as a classifier for classifying the expressions of supplied face into seven basic categories like surprise, neutral, sad, disgust, fear, happy and angry . For face portion segmentation basic image processing operation like morphological dilation, erosion, reconstruction techniques with disk structuring element are used. Six permanent Facial features like eyebrows(left and right), eye (left and right) , mouth and nose are extracted using facial geometry, edge projection analysis and distance measure and feature vector is formed considering height and width of left eye, height and width of left eyebrow, height and width of right eye, height and width of right eyebrow, height and width of nose and height and width of mouth along with distance between left eye and eyebrow, distance between right eye and eyebrow and distance between nose and mouth. Experiments are carried out on JAFFE facial expression database. The paper is organized as follows. Section I gives brief introduction, Section II describes about survey of existing methods, section III highlights on data collection, section IV presents methodology followed, section V gives experimental results and analysis, section VI presents conclusion and future scope and last section gives references used

## II. SURVEY OF EXISTING METHODS

In recent years, the research of developing automatic facial expression recognition systems has attracted a lot of attention. A more recent, complete and detailed overview can be found in [12-14]. Accuracy of facial expression recognition is mainly based on accurate extraction of facial feature components. Facial feature contains three types of information i.e texture, shape and combination of texture and shape information. Feng et. al.[15] used LBP and AAM for finding combination of local feature information , global information and shape information to form a feature vector . They have used nearest neighborhood with weighted chi-sq statistics for expression classification. Feature point localization is done using AAM and centre of eyes and mouth is calculated based on them. Mauricio Hess and G. Martinez [16] used SUSAN algorithm to extract facial features such as eye corners and center, mouth corners and center, chin and cheek border, and nose corner etc. Gengtao zhou et al.[17] used selective feature extraction method where expressions are roughly classified into three kinds according to the deformation of mouth as 1) sad , anger , disgust  2) happy , fear  3) surprise and  again some if then rules are used to sub classify individual group expressions . Jun Ou et al.[18]  used  28 facial feature points and Gabor wavelet filter for facial feature localization , PCA for feature extraction and KNN for expression classification . Md. Zia Uddin , J.J. Lee and T.S. Kim [19] used enhanced Independent component Analysis (EICA ) to extract locally independent component features which are further classified by Fisher linear Discriminant Analysis (FLDA) .Then discrete HMM is used to model different facial Expressions. Feature extraction results of various conventional method ( PCA , PCA-FLDA, ICA and EICA ) in conjunction with same HMM scheme were compared and comparative analysis is presented in terms of recognition  rate . PCA is unsupervised learning method used to extract  useful features and $2^{nd}$ order statistical method for deriving orthogonal bases containing  the maximum variability and is also used for dimensionality reduction .V. Gamathi et al.[20] used uniform local binary pattern (LBP ) histogram technique  for  feature  extraction  and  MANFIS  ( Multiple Adaptive Neuro Fuzzy Inference system) for expression recognition. GRS Murthy and R.S. Jadon[21] proposed modified PCA (eigen spaces) for eigen face reconstruction method for expression recognition. They have divided the training set of Cohn kanade and JAFFE databases into 6 different partitions and eigen space  is constructed for each class , then image is reconstructed.  Mean square error is used as a similarity measure for comparing original and reconstructed image. Hadi Seyedarabi et al.[22] developed facial expression recognition system for recognizing basic expression . They have used cross correlation based optical flow method for extracting facial feature vectors.  RBF neural network and fuzzy inference system is used for recognizing facial expressions. Zhengyou Zhang et al. [23] presented a FER system where they have compared the use of two type of features extracted from face images for recognizing facial expression .Geometric positions of set of fiducial point and multiscale & multi orientation gobor wavelet coefficient extracted from the face image at the fiducial points are the two approaches used for feature extraction. These are given to neural network classifier separately or jointly and results were compared.

Comparison of the recognition performance with different types of features shows that Gabor wavelet coefficients are more powerful than geometric positions. Junhua Li and Li Peng [24] used feature difference matrix for feature extraction and  QNN (Quantum Neural Network) for expression recognition From the survey, it is observed that various approaches have been used to detect facial features [25] and classified as holistic and feature based methods to extract facial feature from images or video sequences of faces. These are geometry based, appearance based, template based and skin color segmentation based approaches. Recently large amount of contributions were proposed in recognizing expressions using dynamic textures features  using both LBP and gabor wavelet approach and appearance features and increases complexity. Moreover one cannot show features located with the help of bounding box. Hence, the proposed facial expression recognition system aimed to use image preprocessing and geometry based techniques for feature extraction and feed forward neural network for expression recognition for the frontal view face images.

### III. DATA COLLECTION

Data required for experimentation is collected from JAFFE database for neural network training and testing. JAFEE stands for The Japanese Female Facial Expression (JAFFE) Database. The database contains 213 images of 7 facial expressions (6 basic facial expressions + 1 neutral) posed by ten different Japanese female models. Sixty Japanese subjects have rated each image on 6 emotion adjectives. The database was planned and assembled by Miyuki Kamachi, Michael Lyons, and Jiro Gyoba with the help of Reiko Kubota as a research assistant. The photos were taken at the Psychology Department in Kyushu University. Few samples are shown in Fig. 1

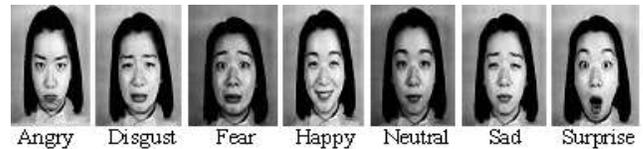
Fig. 1 Few samples of facial expressions of person YM

### IV. METHODOLOGY

Fig. 2 shows the proposed pseudo code for Automatic Facial Expression Recognition System.

1. Read  input image from database and localize face using morphological image processing operations
2. Crop the face image.
3. Extract features from cropped face.
4. Find facial feature vectors.
5. Train neural network.
6. Recognize expression

Fig. 2  Pseudo code for AFERS

#### A. Face Portion Localization and Feature Extraction

Face area and facial feature plays an important role in facial expression recognition. Better the feature extraction rate more is the accuracy of facial expression recognition. Precise localization of the face plays an important role in feature extraction, and expression recognition. But in actual application, because of the difference in facial shape and the quality of the image, it is difficult to locate the facial feature precisely. Images from JAFFE database are taken as input. This database contains low contrast images therefore images are first pre-processed using contrast limited adaptive histogram equalization operation and is used for enhancing contrast of an image. Face area is segmented using morphological image processing operations like dilation, erosion reconstruction, complementation, regional max and clear border(to get Region of Interest).

In order to extract facial features, segmented face image (RoI) is then resized to larger size to make facial components more prominent. SUSAN edge detection operator [26] along with noise filtering operation is applied to locate the edges of various face feature segment components. SUSAN operator places a circular mask around the pixel in question. It then calculates the number of pixels within the circular mask which have similar brightness to the nucleus and refers it as USAN and then subtract USAN size from geometric threshold to produce edge strength image.

Following steps are utilized for facial feature segment localization-

1. Apply Morphological operations to remove smaller segments having all connected components (objects) that have fewer than P pixels where P is some threshold value.
2. Trace the exterior boundaries of segments and draw bounding box by taking into account x,y coordinates and height and width of each segment.
3. Image is partitioned into two regions i.e upper and lower portion on the basis of centre of cropped face assuming the fact that eyes and eyebrows are present in the upper part of face and mouth and nose is present in the lower part. Smaller segments within the region are eliminated by applying appropriate threshold value and remaining number of segments are stored as upper left index, upper right index and lower index. Following criteria is used for selecting appropriate upper left index, upper right index and lower index-
    a) A portion is an upper part if x and y values are less than centx and centy where centx and centy are x- and y-coordinates of center of cropped image. Eyes and eyebrows are present in this area. For left eye and eyebrow portion certain threshold for values of x and y is considered for eliminating outer segments. For right eye and eyebrow also specific threshold value is chosen for eliminating outer segments
    b) A portion is a lower portion if its value is greater than centx and centy where centx and centy are x- and y-coordinates of center of an image. Nose and mouth are present in this area. For nose and mouth area segments, x lies in the specific range and y also lies in certain range is considered as region of interest for eliminating outer segments .Here number of segments for each portion are stored. If number of segments are > 2 then following procedure for combining the segments is called (step 4)
4. Segments are checked in vertical direction. If there is overlapping then the segments are combined. Again if segments are >2 then distance is obtained and the segments which are closer are combined. This process is repeated until we get two segments for each part and in all total six segments(Fig. 4 (j)).

This gives the bounding box for total six segments which will be left and right eyes , left and right eyebrows , nose and mouth features of the supplied face (Fig. 3).

*B. Formation of Feature Vector*

Bounding box location of feature segments obtained in the above step are used to calculate the height and width of left eyebrow, height and width of left eye, height and width of right eyebrow, height and width of right eye, height and width of nose and height and width of mouth. Distance between centre of left eye and eyebrow, right eye and eyebrow and mouth and nose is also calculated. Thus total 15 parameters are obtained and considered as feature vector (Fig.3). Thus-

$$F_v=\{H_1,W_1,H_2,W_2,H_3,W_3,H_4,W_4,H_n,W_n,H_m,W_m,D_1,D_2,D_3\} \quad (1)$$

Where,
$H_1$=height of left eyebrow, $W_1$= width of left eyebrow
$H_2$= height of left eye, $W_2$= width of left eye
$H_3$=height of right eyebrow, $W_3$= width of right eyebrow
$H_4$= height of right eye, $W_4$= width of right eye
$H_n$= height of nose, $W_n$= width of nose,
$H_m$= height of mouth, $W_m$= width of mouth
$D_1$ = distance between centre of left eyebrow and left eye,
$D_2$= distance between centre of right eyebrow and right eye,
$D_3$= distance between centre of nose and mouth

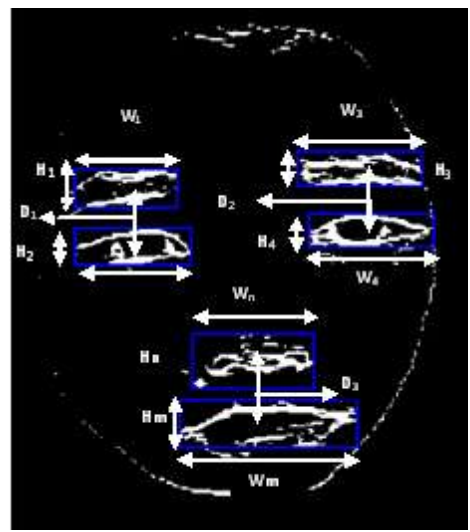

Fig. 3 Feature vector for Expression recognition

## C. Expression Classification using Neural Network

Neural computing has re-emerged as an important programming paradigm that attempts to mimic the functionality of the human brain. This area has been developed to solve demanding pattern processing problems, like speech and image processing. These networks have demonstrated their ability to deliver simple and powerful solutions in areas that for many years have challenged conventional computing approaches. A neural network is represented by weighted interconnections between processing elements (PEs). These weights are the parameters that actually define the non-linear function performed by the neural network. Back-Propagation Networks is most widely used neural network algorithm than other algorithms due to its simplicity, together with its universal approximation capacity. The back-propagation algorithm defines a systematic way to update the synaptic weights of multi-layer perceptron (MLP) networks. The supervised learning is based on the gradient descent method, minimizing the global error on the output layer. The learning algorithm is performed in two stages [27]: feed-forward and feed-backward. In the first phase the inputs are propagated through the layers of processing elements, generating an output pattern in response to the input pattern presented. In the second phase, the errors calculated in the output layer are then back propagated to the hidden layers where the synaptic weights are updated to reduce the error. This learning process is repeated until the output error value, for all patterns in the training set, are below a specified value. The Back Propagation, however, has two major limitations: a very long training process, with problems such as local minima and network paralysis; and the restriction of learning only static input-output mappings [27].

Fifteen values so obtained in the section IV B) are given as an input to the neural network. Model uses an input layer, two hidden layer with 15 and 7 neurons and an output layer.

## V. EXPERIMENTAL RESULTS AND ANALYSIS

In this paper Neural Network model is constructed for JAFEE Face Database for frontal view facial images. Fig. 4 shows the results of facial feature extraction. Initially Face portion segmentation is done using morphological image processing operation and hence face localization is achieved. Region of interest is cropped using localized face and then this image is resized to larger size so that facial feature components should appear prominent. SUSAN edge detection operator along with noise filtering operation is applied to locate the edges of various face feature segment components. Multiple facial feature candidates after applying our algorithm steps 1 and 2 are shown in Fig.4 b).

Cropped Facial image is divided into two regions based on the centre of an image and location of permanent facial feature. The step 3 and 4 are applied to facial segments and results are shown in Fig. 4 (c-i) . Fig 4 j) shows localized permanent facial features which are used as input to neural networks.

**Training Phase:** In this work, supervised learning is used to train the back propagation neural network. The training samples are taken from the JAFFE database. This work has considered 120 training samples for all expressions. After getting the samples, supervised learning is used to train the network. It is trained three times and shown good response in reduction of error signal.

**Testing Phase:** This proposed system is tested with JAFFE database. It was taken totally 30 sample images for all of the facial expressions.

The Fig. 5 shows the GUI for displaying the results of face localization, extracted permanent facial features with bounding box and its recognition and classification. In this figure, the sample image exhibits sad(SA) expression. Performance plot of neural network is shown in Fig. 6. Plot shows that the network learns gradually and reaches towards the goal.

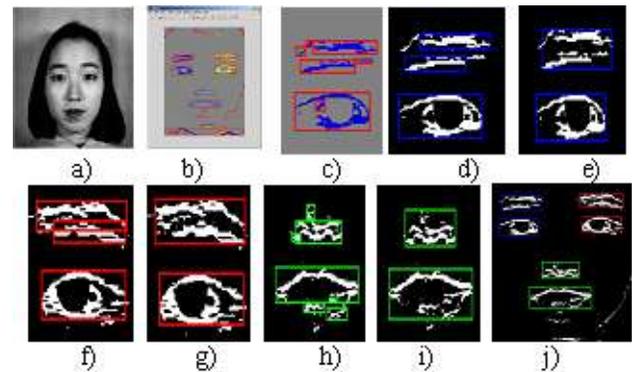

Fig. 4 Results of facial feature extraction a) Original Image b) Multiple facial feature candidates after applying our algorithm step1,2,3 and 4 c) Possible Feature candidates of upper left portion with overlapped eyebrow segments. d) Eye and eyebrow segments after applying step 6 e) Required feature segments of upper left portion f) Possible Feature candidates of upper right portion with overlapped eyebrow segments. g) Required Eye and eyebrow segments after applying step 6 h) Possible Feature candidates of lower portion with overlapped nose and mouth segments. i) Required nose and mouth segments after applying step 6 j) Located Facial features

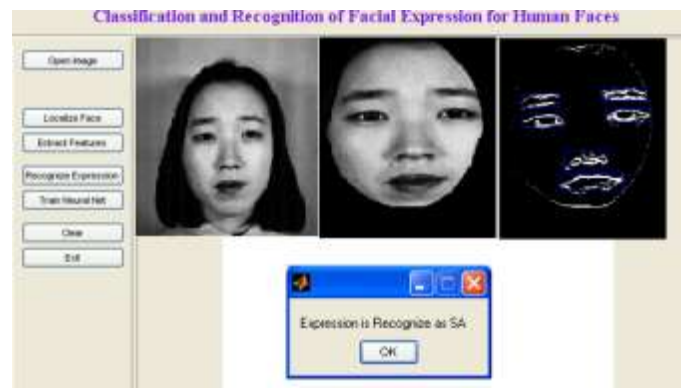

Fig. 5 GUI for classification and recognition of facial

## VI. CONCLUSION AND FUTURE SCOPE

In this paper, automatic facial expression recognition (AFER) system is proposed. Machine recognition of facial expression is a big challenge even if human being recognizes it without any significant delay. The combination of SUSAN edge detector, edge projection analysis and facial geometry distance measure is best combination to locate and extract the facial feature for gray scale images in constrained environments and feed forward back-propagation neural network is used to recognize the facial expression. 100% accuracy is achieved for training sets and 95.26% accuracy is achieved for test sets of JAFFE database which is promising. Table 1 presents the % recognition accuracy of facial expression which appears in literature [28] and our approach. Proposed combination method for feature extraction does not extract exactly six features parameters properly if there are hairs on face area. Therefore in future an attempt can be made to develop hybrid approach for facial feature extraction and recognition accuracy can be further improved using same NN approach and hybrid approach such as ANFIS. An attempt can also be made for recognition of other database images or images captured from camera.

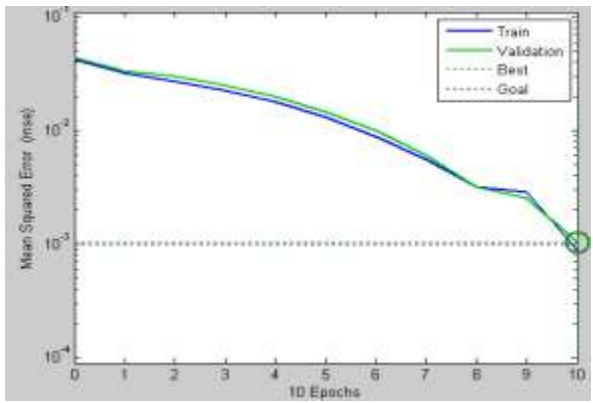

Fig. 4 Performance plot of neural network

TABLE 1
% RECOGNITION ACCURACY

| Authors | No. of subjects Used | Images Tested | % accuracy |
|---|---|---|---|
| Kobayashi and Hara[28] | 15 | 90 | 85 |
| Zhang[12] | 10 | 213 | 90.1 |
| Lyons et. al.[28] | 10 | 193 | 92 |
| Sebe et. al.[28] | - | - | 85-95 |
| Kulkarni SS et. al.[28] | 62 | 282 | 90.4 |
| Chang JY,Chen JL [29] | 08 | 38 | 92.1(for 3 expressions) |
| Our approach | 10 | 30 | 96.42 |

AUTHORS PROFILE

Dr. R.C. Thool is Professor in Deptt. of Information Technology, SGGSIET, SRTMNU, Nanded. His area of interest is computer vision, robot vision and image processing and pattern recognition.

Mrs. S.P. Khandait is a research scholar and Assistant professor in the department of Information Technology, KDKCE, RSTMNU, Nagpur. Presently she is pursuing her PhD in CSE. Her research interest is Image processing and computer vision.

 P. D. Khandait is an Assistant professor in the department of Electronics Engineering , KDKCE, RSTMNU, Nagpur. His area of research interest is Signal and Image processing, soft computing etc..